\documentclass[letterpaper]{article} 
\usepackage{appendix}
\usepackage{aaai24}  
\usepackage{times}  
\usepackage{helvet}  
\usepackage{courier}  
\usepackage[hyphens]{url}  
\usepackage{graphicx} 
\usepackage{natbib}  
\usepackage{caption} 
\usepackage{algorithm}
\usepackage{algorithmic}

\usepackage{newfloat}
\usepackage{listings}
\DeclareCaptionStyle{ruled}{labelfont=normalfont,labelsep=colon,strut=off} 
\floatstyle{ruled}
\newfloat{listing}{tb}{lst}{}
\floatname{listing}{Listing}
\title{Imaginations of WALL-E : Reconstructing Experiences \\with an Imagination-Inspired Module for Advanced AI Systems}
\author{
    Zeinab Taghavi\textsuperscript{\rm 1},
    Soroush Gooran\textsuperscript{\rm 2},
    Seyed Arshan Dalili\textsuperscript{\rm 3},
    Hamidreza Amirzadeh\textsuperscript{\rm 4},
    Mohammad Jalal Nematbakhsh\textsuperscript{\rm 5},
    Hossein Sameti\textsuperscript{\rm 6}
}

\affiliations{
    \textsuperscript{\rm 1, \rm 2, \rm 3, \rm 4, \rm 5, \rm 6}Sharif University of Technology\\
    zeinabtaghavi@sharif.edu, gooran@sharif.edu, adalili@ce.sharif.edu, hamid.amirzadeh78@sharif.edu, jalal.nematbakhsh99@sharif.edu, sameti@sharif.edu
}

\usepackage{bibentry}

\begin{document}

\maketitle

\begin{abstract}

    In this paper, we introduce a novel Artificial Intelligence (AI) system inspired by the philosophical and psychoanalytical concept of imagination as a ``Re-construction of Experiences". Our AI system is equipped with an imagination-inspired module that bridges the gap between textual inputs and other modalities, enriching the derived information based on previously learned experiences. A unique feature of our system is its ability to formulate independent perceptions of inputs. This leads to unique interpretations of a concept that may differ from human interpretations but are equally valid, a phenomenon we term as ``Interpretable Misunderstanding". We employ large-scale models, specifically a Multimodal Large Language Model (MLLM), enabling our proposed system to extract meaningful information across modalities while primarily remaining unimodal. We evaluated our system against other large language models across multiple tasks, including emotion recognition and question-answering, using a zero-shot methodology to ensure an unbiased scenario that may happen by fine-tuning. Significantly, our system outperformed the best Large Language Models (LLM) on the MELD, IEMOCAP, and CoQA datasets, achieving Weighted F1 (WF1) scores of 46.74\%, 25.23\%, and Overall F1 (OF1) score of 17\%, respectively, compared to 22.89\%, 12.28\%, and 7\% from the well-performing LLM. The goal is to go beyond the statistical view of language processing and tie it to human concepts such as philosophy and psychoanalysis. This work represents a significant advancement in the development of imagination-inspired AI systems, opening new possibilities for AI to generate deep and interpretable information across modalities, thereby enhancing human-AI interaction.

\end{abstract}

\section{Introduction}

\begin{figure}[t]
\centering
\includegraphics[width=0.5\textwidth]{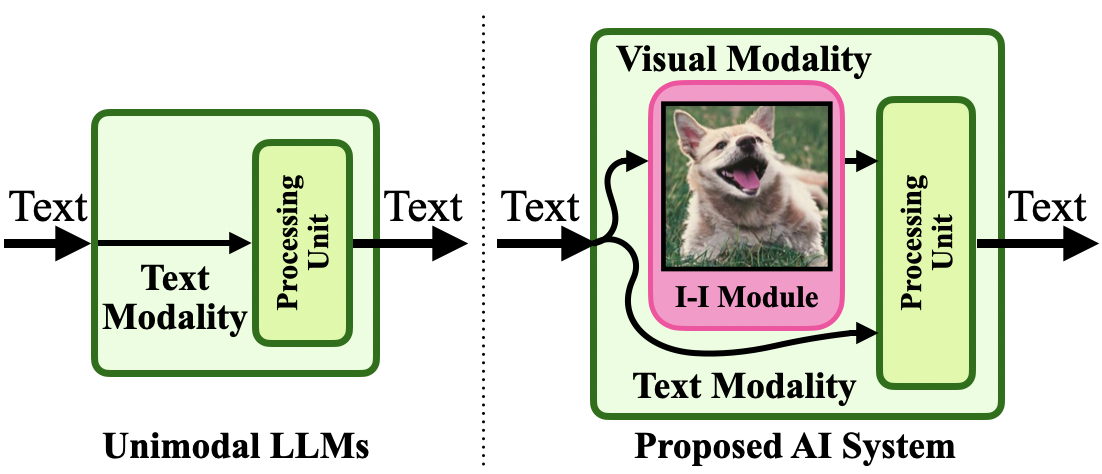} 
\caption{\textit{Proposed system in comparison to Unimodal LLMs in textual tasks.} As you can see, the human-interpretable input for processing unit in common LLMs is just input text, but in our proposed system, the system generates human-interpretable input by itself in another modality for MLLM. This process happens in the Imagination-Inspired (I-I) module.}
\label{our_sys}
\end{figure}

\begin{figure}[t]
\centering
\includegraphics[width=0.25\textwidth]{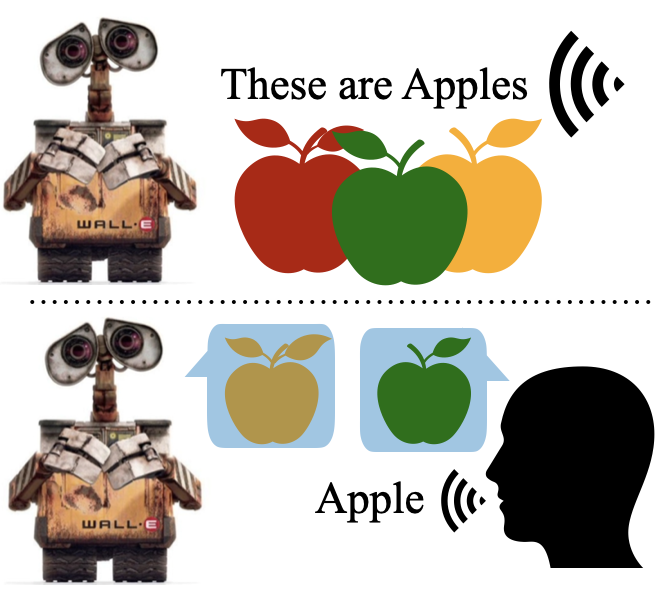} 
\caption{\textit{``Re-construction of Experiences".} When we reconstruct information based on what we have seen before. It might be a bit different, but same in major points which were mentioned in the input. This can cause ``Interpretable Misunderstanding".}
\label{reconstrcution}
\end{figure}

Imagination, as defined in the Cambridge dictionary, is ``the ability to form pictures in the mind" \cite{cambridge_online_dict}. In some philosophical papers, it is considered as the process of ``Re-construction of Experiences" \cite{Currie2002_CURRMI, Van_Leeuwen_2013}. It allows us to form mental images from text, drawing on our past experiences like what happens in Figure \ref{reconstrcution}. These mental images substantially influence our understanding and emotional reactions to various concepts \cite{Van_Leeuwen_2014, Paivio1990-ac}. This theory, emphasized by psychoanalysts, highlights the crucial interplay between past experiences and current perception \cite{Freud2004-gc, Lacan2007-mi}.

For the first time, drawing inspiration from this philosophical and psychoanalytical foundation, we introduce an AI system embedded with an imagination-inspired module, as you can see in Figure \ref{our_sys}. This system bridges the gap between textual inputs and other modalities, thereby enriching the information derived from the text, using previously learned experiences which act as ``Re-Construction of Experiences". A unique attribute of this AI system is its capability to formulate its own independent perception of an input. Thus, Our system can arrive at unique, yet equally valid, interpretations of a concept—a phenomenon we term as AI's ``Interpretable Misunderstanding". For example, when we give the system input ``apple", we may have in mind a green apple. In contrast, the image generated by the AI system may depict a yellow apple. Nevertheless, both representations are still considered as apples. Furthermore, the generated image may contain additional information, such as being curved or having leaves, which were not mentioned directly. This is not a computational mistake, this is a ``Different Point of View" and it should be considered important, as you can see in Figure \ref{reconstrcution}. 

For adding this imagination-inspired module to our system for a unimodal textual task (text-input text-output), we made a pipeline which is described in Figure \ref{our_sys}. In our implementation, our system takes a text as input, then generates an image using a text-to-image model \cite{stable_diff_v2}. Due to the inherent randomness of text-to-image models, the generated image varies each time, even for the same text input \cite{stable_diff_v2}. Finally, by giving both text and image to an MLLM \cite{instructblip}, the system provides us with the text output. Generating meaningful information from textual input to other modalities, which in this case is image, roles as the imagination-inspired module. While fine-tuning often enables models to excel in specific tasks \cite{Devlin_Chang_Lee_Toutanova_2018, liu2019roberta, chowdhery2022palm, touvron2023llama}, we shifted our focus towards Large-scale Language Models for their depth and interactivity. Our system prominently features an MLLM at its core. This design choice not only enables our system to extract valuable insights from different modalities but also ensures its primary operation remains unimodal. Additionally, the model's substantial size allows the integration of an imagination-inspired module as a secondary smaller component behind the primary bigger language processing module. Using Large-scale Language Models ensures the model's versatility across multiple tasks without the need for fine-tuning.
\\
We assessed our system by benchmarking its performance against other LLMs across multiple tasks. We initially concentrated on emotion recognition, using the MELD and IEMOCAP datasets \cite{MELD, IEMOCAP}. Subsequently, we focused on the question-answering task using the CoQA dataset \cite{CoQA}. The CoQA dataset aligns with our aim to evaluate the model's ability to comprehend the text based solely on the provided input without relying on any pre-existing knowledge. More reasons are in Appendix 1.
\\
We compared our system to other LLMs exhibiting robust performance on the TriviaQA dataset \cite{triviaqa}. Our approach employs a zero-shot methodology, which allows us to assess the models' raw abilities without task-specific fine-tuning, ensuring an unbiased evaluation scenario. Our built system outperformed the best LLM (Koala-7B) on the MELD, IEMOCAP, and CoQA datasets. This accomplishment signals a remarkable advancement in developing AI systems by adding the imagination-inspired module.
\\
Our work paves the way for a new era in artificial intelligence, where AI systems can generate deep and meaningful information from input in one modality, starting with images and potentially expanding to audio or even action in the future. Imagine a day when we tell an AI system that we're walking, and it can reconstruct the experience within itself, thereby enhancing human-AI interaction. 

\section{Related Works}

Here we will delve into the current state of LLMs and MLLMs in AI, the philosophical and psychoanalytical underpinnings of imagination in order to discuss our system's implementation.

\subsection{Current State of LLMs and MLLMs in AI}
In recent years, Large Language Models (LLMs) have emerged as a transformative technology in natural language processing, exhibiting remarkable capabilities in various tasks such as text generation and question-answering.

Subsequently, Unimodal-trained LLMs, such as GPT-3 \cite{Brown_Mann_Ryder_Subbiah_Kaplan_Dhariwal_Neelakantan_Shyam_Sastry_Askell_et_al_2020}  and BERT \cite{Devlin_Chang_Lee_Toutanova_2018}, leveraged transformer models, delivering significant improvements in language understanding tasks but remaining confined to the linguistic modality \cite{hoffmann2022training}.

The AI field then progressed towards multimodal models, capable of processing diverse data types \cite{yin2023survey, kiros2014multimodal}. However, these models typically relied on user-provided multimodal data \cite{multimodal2011ng, gong2023multimodalgpt, zhu2023minigpt4, driess2023palme, zhang2023transfer}.

Unlike previous works, our study adopts a modality-conversion approach \cite{Change_of_Heart}, wherein the AI system generates additional modality data, eliminating the exclusive dependence on user-provided data. This enables the AI system to process, generate, and integrate multiple data types, creating a more comprehensive representation of the problem. The effectiveness of this innovative approach is demonstrated through our experimental results.

\subsection{The Essence of Imagination in Philosophy}

Constructive imagination is a complex cognitive process that goes beyond mere whimsical thought. It forms a fundamental building block of our cognitive and perceptual experiences and plays a crucial role in our understanding of the world \cite{Van_Leeuwen_2013, Van_Leeuwen_2014}. In the context of our AI system, understanding the philosophy of constructive imagination is essential. This section explores key elements of constructive imagination that have informed the design of our system.

\begin{itemize}

\item \textbf{Connection to Sensory Perception:} Constructive imagination engages our ability to amalgamate elements from memory, sensory experiences, and emotions \cite{Van_Leeuwen_2013}. Brain studies indicate that the parts of the brain responsible for sensory perception also play a crucial role in the imagination process. These areas activate not only during sensory experiences but also when visualizing content related to a specific sensory modality, showing the importance of sensory parts of the brain in imagination \cite{Slotnick_Thompson_Kosslyn_2005}.

\item \textbf{Influence of Beliefs:} Our beliefs, largely shaped by our past experiences, exert considerable influence over our imaginative processes. Studies show how people's beliefs influence their understanding of stories \cite{Weisberg_Goodstein_2009,Harris_2000}. This suggests that initiating imagination depends heavily on what we believe, meaning past beliefs affect current imagination \cite{Nichols_2000,Nichols_Stich_2003}.

\item \textbf{Generation of New Ideas:} In constructive imagination, we can generate new imaginative ideas based on existing ones. For example, when we learn about magic or spaceships, we use our beliefs to further elaborate on initial imaginings \cite{Van_Leeuwen_2013, Lillard_Witherington_2004}

\item \textbf{Influence on Actions and Emotions:} It's important to note that imaginative representations can directly influence actions and elicit emotional responses \cite{Funkhouser2009_FUNIAO_2,Nanay2013_NANBPA,Nichols_2006}. This connection between imagination and action underscores the potential applications of our AI system in various fields.

\end{itemize}

In summary, the philosophy of constructive imagination offers valuable insights into the cognitive processes that underlie imagination. These insights have been instrumental in shaping our AI system, enabling it to bridge the gap between textual inputs and other modalities, and generating deep and meaningful information.

\subsection{AI Through the Psychoanalytic Lens}

AI has been making strides in mimicking human cognitive processes, and one area of human cognition that offers intriguing possibilities is psychoanalysis. Psychoanalysis, a field of psychology that delves into the unconscious mind, could provide valuable insights into the development of AI systems. This section explores how the principles of psychoanalysis can be applied to AI, enhancing its ability to interpret and respond to inputs in a way that more closely resembles human cognition.

Psychoanalysis, founded by Sigmund Freud, posits that the unconscious mind houses memories, desires, and feelings that may influence behavior and thought processes without conscious awareness \cite{Freud2004-gc, Lacan2007-mi}. Freud's view provides a useful framework for understanding how past experiences can influence current behavior. For instance, a suppressed memory from childhood, such as a fear of darkness, can manifest in an adult's preference for well-lit environments, influencing their behavior without them being consciously aware of the cause \cite{Freud2004-gc}.

Lacan, another influential figure in psychoanalysis, offered another perspective on the unconscious mind. He likened it to ``spoken language", composed of symbols, images, and patterns of speech that unknowingly shape individuals. This suggests that the unconscious mind is not just shaped by our social interactions and language use but also plays an active role in forming our individuals \cite{Lacan2007-mi}.

It is suggested that future AI systems could possess aspects of human psychological constructs, such as consciousness and unconsciousness \cite{Wallace-Consciousness, Kurzweil2005-ca}. These psychoanalytical theories offer a roadmap for developing systems that can generate a nuanced understanding of information, much like how humans use past experiences to inform present understanding. By incorporating elements of psychoanalysis into the design of our AI system, we aim to create a system that can interpret and respond to inputs in a way that more closely resembles human cognition. 

\begin{figure}[t]
\centering
\includegraphics[width=0.4\textwidth]{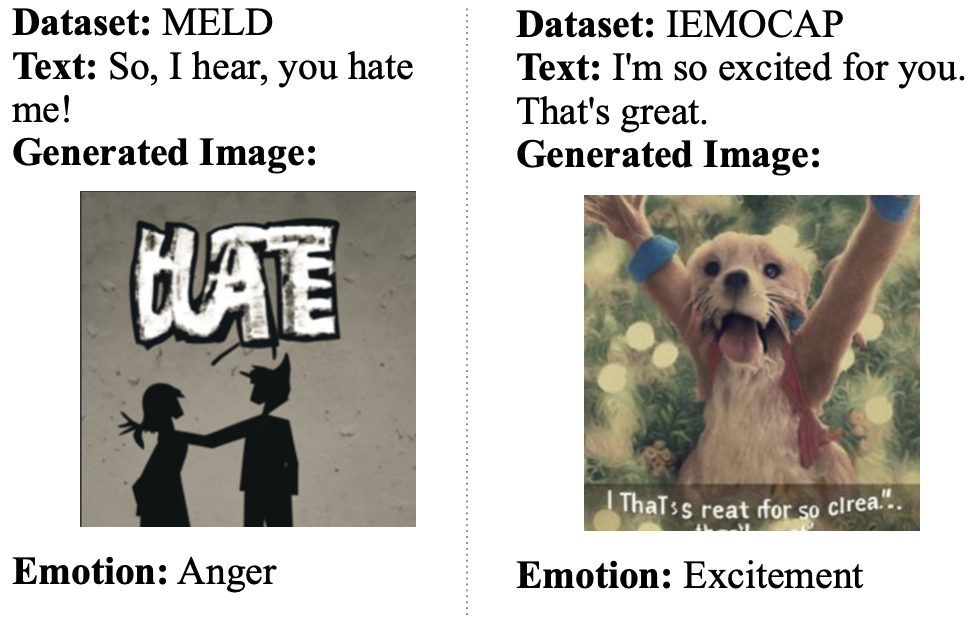} 
\caption{\textit{MELD and IEMOCAP Text-to-Image sample.} The Text-to-Image model was utilized to convert the input sequence of the datasets into a visual representation, resulting in an image. Finally, this image is served as input to the MLLM. True emotion is provided in below of the image.}
\label{DS_sample_MELD_IEMOCCAP}
\end{figure}

\begin{figure*}[t]
\centering
\includegraphics[width=0.8\textwidth]{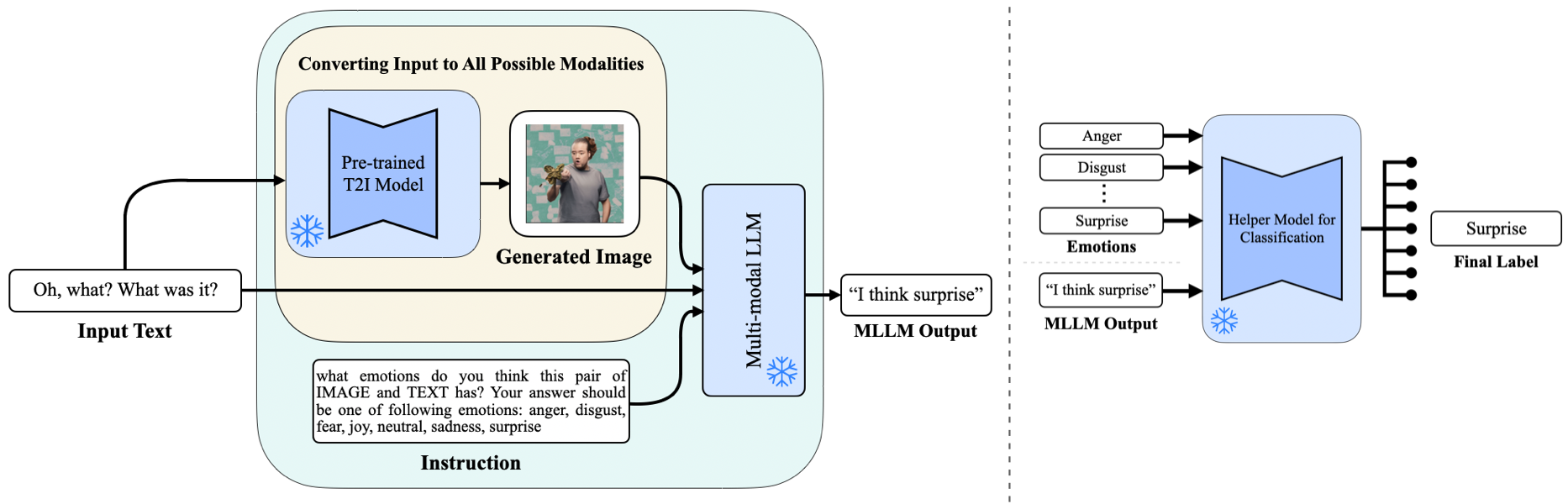} 
\caption{\textit{Emotion recognition system.} The Text-to-Image model translates input text into an image and forwards it to the MLLM. A helper model is then used to identify the label with the highest cosine similarity score between the MLLM's output and the emotion embeddings.}
\label{system_arch_ER}
\end{figure*}

\begin{figure}[t]
\centering
\includegraphics[width=0.4\textwidth]{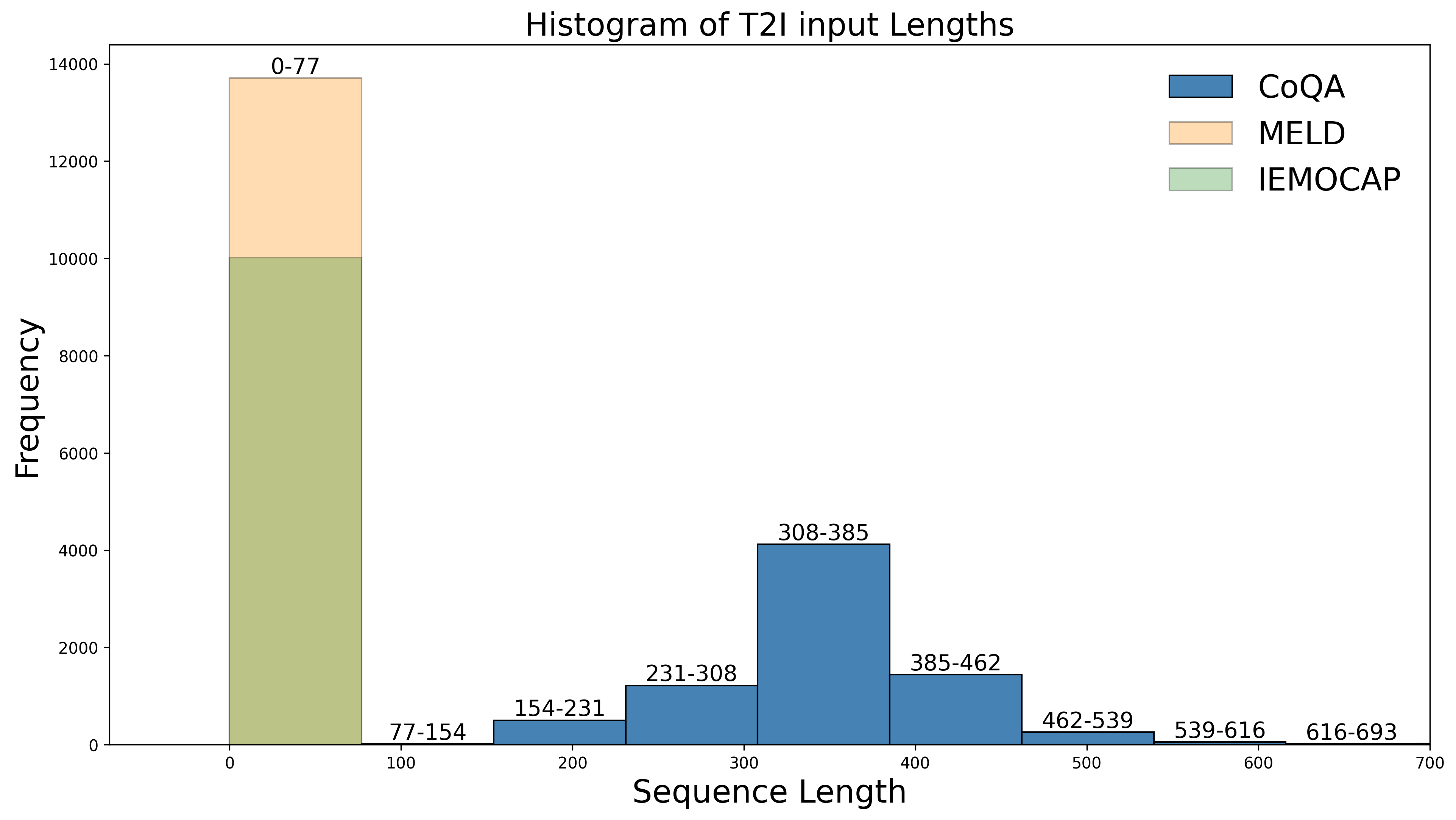} 
\caption{\textit{Histogram of the Text-to-Image model input lengths.} All input sequences from the MELD and IEMOCAP datasets have a length of fewer than 77 tokens, which aligns with the maximum sequence length of the Stable Diffusion tokenizer. However, the input stories in the CoQA dataset are longer, necessitating the need for segmenting the stories.}
\label{Histogram_of_T2I_Input_Lengths}
\end{figure}

\begin{figure*}[t]
\centering
\includegraphics[width=0.6\textwidth]{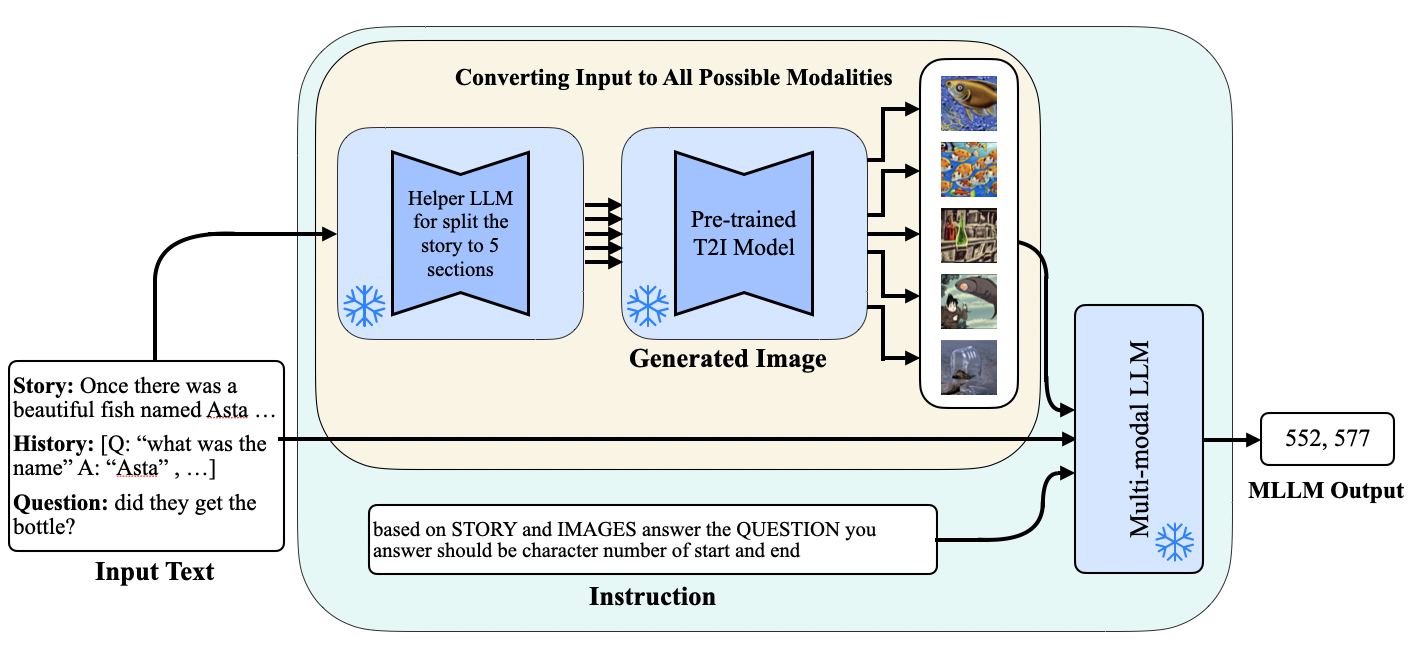} 
\caption{\textit{Question answering system.} A helper LLM converts the story to five segments, then the Text-to-Image model converts each segment into an image and feeds into the MLLM model, along with the question and prior responses, to generate a span of the answer to the question.}
\label{system_arch_QA}
\end{figure*}

\begin{figure*}[t]
\centering
\includegraphics[width=0.75\textwidth]{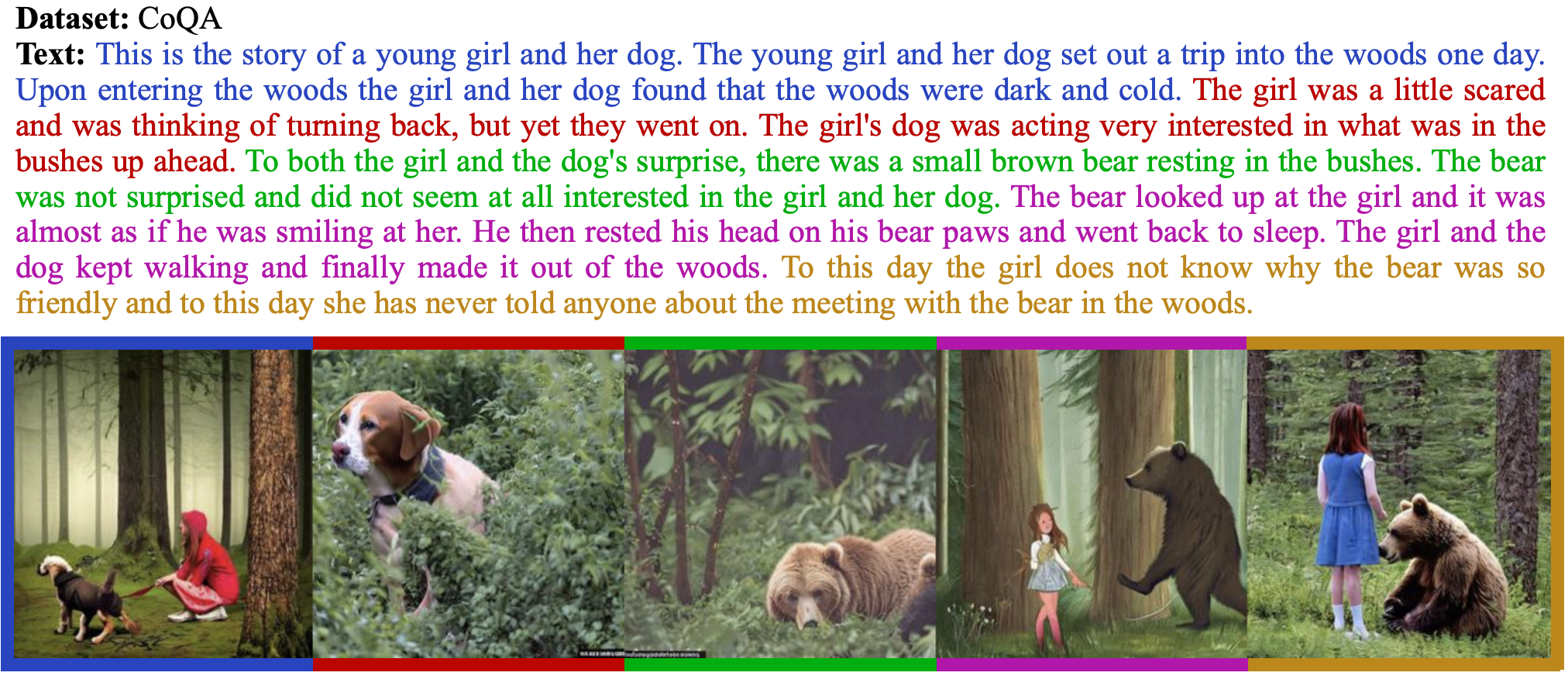} 
\caption{\textit{CoQA Text-to-Image sample.} GPT3.5 as a helper LLM was employed to partition each narrative into five distinct sections. Subsequently, the Text-to-Image model was utilized to convert each section into a visual representation, resulting in a set of five images. Finally, these images were concatenated horizontally to form a single composite image, which served as input to the MLLM. Each distinct section is in a different color and the same as the corresponding generated image's borderline. }
\label{DS_SAMPLE_CoQA}
\end{figure*} 

\begin{table*}[t]
    \centering
    \begin{tabular}{|c|c|c|c|c|c|}
        \hline
        Model & Modality & Image & Text & Instruction & Output Processing \\
        \hline
        Gen\_Image\_Inp\_Text\_Both & Multimodal & Generated  &  Input & Consider Inputs & - \\
        Gen\_Image\_Inp\_Text\_Txt & Multimodal & Generated  & Input & Consider Text & - \\
        Gen\_Image\_Inp\_Text\_Img & Multimodal & Generated  & Input & Consider Image & - \\
        Gen\_Image\_No\_Text\_Img & Multimodal & Generated  & None & Consider Image & - \\
        Gen\_Image\_Inp\_Text\_P1 & Multimodal & Generated  & Input & Choose one of emotions & - \\
        v\_Image\_Inp\_Text\_P2 & Multimodal & Generated  & Input & What emotion does perceive & - \\
        Gen\_Image\_Inp\_Text\_P3 & Multimodal & Generated  & Input & No Instruction & - \\
        Dem\_Image\_Inp\_Text\_Both & Multimodal & Demo & Input & Consider Inputs & - \\
        \hline
        MPT-7B & Unimodal & - & Input & - & Yes \\
        \hline
        GPT-J-6B & Unimodal & - & Input & - & Yes \\
        \hline
        Pythia-6.9b & Unimodal & - & Input & - & Yes \\
        \hline
        Cerebras-GPT-2.7B & Unimodal & - & Input & - & Yes \\
        \hline
        StableLM-7B & Unimodal & - & Input & - & Yes \\
        \hline
        Koala-7B & Unimodal & - & Input & - & Yes \\
        \hline
        OPT-6.7B & Unimodal & - & Input & - & Yes \\
        \hline
    \end{tabular}
    \caption{\textit{Experiments guide.} This Table presents the experimental specifications for each investigation, including the type of LLM utilized: Multimodal or Unimodal; How the images are obtained: Generated (Gen), Demo (Dem) or None (-); The inputs provided by the user: Input text (Inp)  or None (No); The instruction direction employed: Both Inputs (Both), Text (Txt), Image (Img) or Special Experiments Instruction (P1, P2, and P3); Finally whether any output processing was also conducted (Yes) or Not (-).}
    \label{experiments}
\end{table*}

\begin{table*}[t]
    \centering
    \begin{tabular}{|c|c|c|cc|cc|}
        \hline
        & & & \multicolumn{2}{c|}{\textbf{MELD}} & \multicolumn{2}{c|}{\textbf{IEMOCAP}} \\
        \cline{4-7}
        \textbf{Experiments} & \textbf{Modality} &  \textbf{Output Processing} & \textbf{WF1(\%)} & \textbf{Acc(\%)} & \textbf{WF1(\%)} & \textbf{Acc(\%)} \\
        \hline
        Gen\_Image\_Inp\_Text\_Both &  &  - & 40.13 & 40.76 & \textbf{25.23} & \textbf{25.79} \\
        Gen\_Image\_Inp\_Text\_Tex &  &  - & \textbf{46.74} & \textbf{46.51} & 23.25 & 25.09 \\
        Gen\_Image\_Inp\_Text\_Img &  &  - & 42.16 & 42.29 & 22.35 & 24.25 \\
        Gen\_Image\_No\_Text\_Img &  &  - & 29.60 & 30.72 & 6.92 & 15.63 \\
        Gen\_Image\_Inp\_Text\_P1 & Multimodal &  - & 45.19 & 44.98 & 22.55 & 24.50 \\
        Gen\_Image\_Inp\_Text\_P2 &  &  - & 40.96 & 39.27 & 12.56 & 13.59 \\
        Gen\_Image\_Inp\_Text\_P3 &  &  - & 14.90 & 12.03 & 10.54 & 8.46 \\
        Dem\_Image\_Inp\_Text\_Both &  &  - & 43.61 & 43.56 & 24.3 & 25.59 \\
        \hline
        MPT-7B & Unimodal &  Yes & 10.09 & 16.24  & 11.16 & 10.70 \\
        \hline
        MPT-7B & Unimodal &  - & 5.73 & 15.36 & 10.51 & 12.35 \\
        \hline
        GPT-J-6B & Unimodal &  Yes  & 18.71 & 17.31 & 8.44 & 8.21 \\
        \hline
        GPT-J-6B & Unimodal &  - & 7.78 & 16.09 & 4.63 & 9.56 \\
        \hline
        Pythia-6.9B & Unimodal &  Yes & 5.23 & 15.13 & 3.83 & 11.15 \\
        \hline
        Pythia-6.9B & Unimodal &  - & 5.48 & 15.40 & 3.57 & 11.55 \\
        \hline
        Cerebras-GPT-2.7B & Unimodal &  Yes & 15.09 & 13.67 & 8.51 & 8.01 \\
        \hline
        Cerebras-GPT-2.7B & Unimodal &  - & 7.80 & 15.59 & 5.09 & 9.66 \\
        \hline
        StableLM-7B & Unimodal &  Yes & 22.89 & 20.91 & 10.20 & 8.96 \\
        \hline
        StableLM-7B & Unimodal &  - & 7.75 & 16.13 & 5.12 & 9.06 \\
        \hline
        Koala-7B & Unimodal &  Yes & \textbf{22.89} & 19.42 & \textbf{12.28} & \textbf{13.34} \\
        \hline
        Koala-7B & Unimodal &  - & 10.76  & \textbf{35.24} & 7.42 & 11.40 \\
        \hline
        OPT-6.7B & Unimodal &  Yes & 5.59 & 15.28 & 3.79 & 11.25 \\
        \hline
        OPT-6.7B & Unimodal &  - & 5.5 & 15.47 & 3.8 & 11.40 \\
        \hline
    \end{tabular}
    \caption{Weighted F1 and Accuracy scores of multimodal and unimodal LLMs for emotion recognition task on MELD and IEMOCAP datasets.}
    \label{ER_results}
\end{table*}

\begin{table}[t]
    \centering
    \begin{tabular}{|c|c|c|c|}
    	\hline
    	\textbf{Experiments} & \textbf{Modality} & \textbf{OF1 (\%)}\\ 
    	\hline	
    	Gen\_Image\_Inp\_Text\_Both &  & 17.0\\
    	Gen\_Image\_Inp\_Text\_Txt &  & 16.7\\
    	Gen\_Image\_Inp\_Text\_Img & Multimodal & 15.3\\
    	Gen\_Image\_No\_Text\_Img &  & 1.7\\
    	Dem\_Image\_Inp\_Text\_Both &  & 16.7\\
    	\hline
    	Koala-7B & Unimodal &  2.1 \\
    	StableLM-7B & Unimodal & 1.1  \\
    	MPT-7B & Unimodal &  1.9  \\
    	Cerebras-GPT-2.7B & Unimodal &  7.0  \\	
    	\hline
    \end{tabular}
    \caption{Overall F1 (OF1) scores of multimodal and unimodal LLMs for question-answering task on the CoQA dataset.}
    \label{QA_results}
\end{table}

\section{The Proposed System: AI Systems with Imagination-Inspired Module}

In the field of artificial intelligence, we often utilize datasets that pair texts with images under the assumption that these pairs are accurately matched \cite{coco, LAION_5B}. We can train a model to generate an image from a piece of text. The objective is to discover a pattern or distribution that, when sampled, yields text and image pairs akin to those in our original dataset. This objective is encapsulated in the model's ``loss function", which quantifies the discrepancy between the model's predictions and the actual data.

Our system operates by taking a piece of text and generating an image based on the pattern it has learned from the correctly matched examples in the dataset \cite{stable_diff_v1,stable_diff_v2}. This learned pattern serves as a memory bank for the model, storing the information it has gleaned from past data. This stored information, which we can think of as the model's 'experiences', enables the model to generate an image from text. 

In the aspect of philosophy, this process mirrors the human cognitive process of \textbf{``Re-construction of Experiences"} based on a given text, as discussed in the articles \cite{Van_Leeuwen_2013, Van_Leeuwen_2014}.
In the aspect of psychoanalysis, these systems, trained on historical data, generate an image similar to the one they were trained on when presented with a familiar text. This can influence the model's current perception in ways that \textbf{may not be immediately apparent}. For instance, if our model is trained with images associated with the concept of fear, and these images are predominantly dark, the system is likely to generate a dark image when given a text with the emotion of fear. This behavior mirrors the way our unconscious mind influences our perceptions and actions based on past experiences.

We employ text-to-image models to augment the input data, enabling us to ``Re-construction of Experiences", a process we refer to as drawing ``Inspiration from Imagination". We incorporate this as an imagination-inspired module in our system. The output of this module, along with the user's input, is then given the MLLM by the specific instruction, which is determined based on the task and experiment. 

\subsection{System Architecture}

The proposed system architecture designed for tasks involving image generation from textual inputs and leveraging MLLMs for zero-shot emotion recognition and question answering. The foundation of our system is common for both tasks, and two distinct architectures were required to accommodate the unique data features of each task.

\subsubsection{Common Foundation}

The common foundation of our system involves generating an image from the input text using text-to-image models. After that, the input text and generated image are passed to an MLLM, as you can see in Figure \ref{our_sys}. We employed Stable Diffusion V2 for the text-to-image system and Instructblip as our MLLM due to its capability to interpret instructions \cite{stable_diff_v2, instructblip}.

\subsubsection{Emotion Recognition Architecture}

For emotion recognition, the system processes the input text and generates an image, as you can see in Figure \ref{DS_sample_MELD_IEMOCCAP}. Both the text and image are then forwarded to the MLLM. The model is guided in selecting an emotion through a series of instructions. However, the outputs from the MLLM and LLM may exceed the intended output label. To address this, we employ a helper model. This model embeds the output of the MLLM (or LMM) and each emotion, then identifies the label with the highest cosine similarity score, as you can see in Figure \ref{system_arch_ER} \cite{thakur-2020-AugSBERT}. Sometimes the output of LLMs starts with the input given to the model, followed by the answer produced by the model. In these cases, we separate the answer field and examine the new results named ``output-processing". Examples are provided in Appendix 2.

\subsubsection{Question-Answering Architecture}

For the question-answering task, special adjustments were necessary due to the longer textual inputs of the CoQA dataset. As shown in the token histogram in Figure \ref{Histogram_of_T2I_Input_Lengths}, most stories had between 308 and 385 tokens, which is five times the capacity of the Stable Diffusion 2's tokenize. To address this, stories were divided into five sections using GPT-3.5. We asked GPT-3, as the helper model, to give us four token numbers to split the story. The story was divided at the nearest full stop to the token numbers specified. These sections were then converted into images and concatenated horizontally for input to the MLLM, as shown in Figure \ref{DS_SAMPLE_CoQA}. For each question, the model was provided with the story, previous questions, and prior responses to generate the final answer, as you can see in Figure \ref{system_arch_QA}.

In summary, the proposed system architecture is tailored to the specific requirements of emotion recognition and question answering, leveraging text-to-image models and MLLMs to achieve nuanced understanding and response generation.

\section{Experiments and Results}

We had some experiments in the direction of some questions. To conduct experiments, we utilized various instructions to guide the model's attention to specific parts of the input. Implementation tips are provided in Appendix 3. As shown in the Table \ref{experiments}, our approach involved these:

\begin{itemize}
    \item Instructed to pay attention to both the textual and visual inputs (Gen\_Image\_Inp\_Text\_Both).
    \item Instructed to focus only on the text input, despite receiving both textual and visual inputs (Gen\_Image\_Inp\_Text\_Txt).
    \item Instructed to consider only the image while still receiving both textual and visual inputs (Gen\_Image\_Inp\_Text\_Img).
    \item Instructed to consider only the image while receiving no textual input and just processed visual inputs (Gen\_Image\_No\_Text\_Img).
    \item Special instruction 1: instructed to look at the task as a classification task (not emotion recognition), but the dataset's labels were given to the model by receiving both textual and visual inputs (Gen\_Image\_Inp\_Text\_P1).
    \item Special instruction 2: instructed to choose an emotion (no label was given, and the model was free to choose any emotion) by receiving both textual and visual inputs (Gen\_Image\_Inp\_Text\_P2).
    \item Special instruction 3: No instruction was given to the model, and it was free to generate any text by receiving both textual and visual inputs (Gen\_Image\_Inp\_Text\_P3).
    \item Instructed to pay attention to both the textual and visual inputs. But the image was a Demo image (Dem\_Image\_Inp\_Text\_B).
\end{itemize}

We choose LLMs with a good score on the TriviaQA dataset, including MPT \cite{MosaicML2023Introducing}, GPT-J \cite{gpt-j}, Pythia \cite{biderman2023pythia}, Cerebras-GPT \cite{dey2023cerebrasgpt}, StableLM \cite{gpt-neox-library}, Koala \cite{koala_blogpost_2023} and OPT \cite{zhang2022opt}.

The results are in Tables \ref{ER_results} \ref{QA_results} and the questions are as follows:
Due to being a classification task in emotion recognition, the metric is the Weighted F1 score (WF1), and the metric for the question-answering task is the Overall F1 Score (OF1).

\subsubsection{Is Adding of the Imagination-inspired Module Effective?}

In addition to the good performance of LMMs in several tasks, in both emotion recognition and question-answering tasks, in our experiments, the results were that our proposed system was able to significantly (WF1: 46.74\%, 25.23\%, and OF1:17\% for MELD, IEMOCAP and CoQA datasets) outperforms the best model among LLMs (Koala-7B model, with WF1: 22.89\%, 12.28\% for MELD and IEMOCAP, and Cerebras-GPT-2.7B with WF1:7\% for the CoQA dataset), which shows performance improvement by using imagination-inspired module in our system.

\subsubsection{Is Multimodality Sufficient, or Does Independent Perception Through Generated Images Truly Make a Difference?}

We conducted an experiment using a static image provided as a demo on the official MLLM GitHub page instead of the generated image. The results (WF1: 43.61\%, 24.3\%, and OF1:16.7\% for MELD, IEMOCAP, and CoQA datasets) outperformed the best model among the unimodal-trained LLMs (Koala-7B  model, with WF1: 22.89\%, 12.28\%, and Cerebras-GPT-2.7B model, with OF1: 7\% for MELD, IEMOCAP and CoQA datasets). That clearly shows the ability of multimodal approaches and their superiority of them over unimodal approaches, even for inherently unimodal tasks.
\\
However, the best results achieved using generated images surpassed those achieved with the demo image and LLM (WF1: 46.74\%, 25.23\%, and OF1:17\% for MELD, IEMOCAP, and CoQA datasets). Which again shows the potential of using imagination-inspired modules.

\subsubsection{Can only Independent Perception Through Generated Images be in line with emotions?}
Our results unveiled significant potential for independent perception through generated images. Our MLLM exhibited superior performance over the leading unimodal model, Koala-7B, registering a WF1 score of 29.60\% compared to Koala-7B's 22.89\% on the MELD dataset. Although this outperformance was not consistent across all datasets, it underscored the impressive capabilities of our system in processing and understanding emotions, even in the absence of textual data. That also illuminated the potential of processing only generated images, interpreted as ``the system's imagination", in conveying information.

\subsubsection{Step-by-Step Instruction: Do the Model's Outputs Align with a Specific Emotion without Explicit Emotion Detection Instruction?}

We evaluated the use of ``Gradually Introduced" instructions. Initially, we examined the model's outputs without any specific instructions (P3). Subsequently, we tested the instruction like ``What emotion do you perceive in this sentence" (P2). After That, the instruction ``This is a classification task, choose one of the emotions" followed by the dataset's emotions (P1). At the End of our primary test, the complete instruction ``What emotions do you think this pair of IMAGE and TEXT has?" along with the dataset emotions (Gen\_Image\_Inp\_Text\_Both). We gave text and images to the model in all these four experiments. Exact instructions are provided in Appendix 2.
\\
We observed improvement of quality in each step from P3 to P1 (WF1: 14.90\%, 40.96\%, and 45.19\% for MELD dataset, WF1: 10.54\%, 12.56\% and 22.55\% for IEMOCAP dataset). The results highlight the crucial role of instruction in enhancing the model's performance.
\\ 
By mapping MLLM's outputs to labels in the P3 test using helper model, our system outperformed (WF1:14.9\% and 10.54\% for MELD and IEMOCAP datasets) some LLMs such as MPT-7B (WF1:10.09\% for MELD dataset), Pythia-6.9b (WF1: 5.48\% and 3.83\%  for MELD and IEMOCAP datasets), and OPT-6.7B  (WF1:5.59\% and 3.8\% for MELD and IEMOCAP datasets). So we can say the system's output tended to reflect the emotional signs conveyed from the input. This finding suggests that our system can effectively perceive and express emotions in its output, even without explicit guidance. For instance, when the input text is filled with anger or happiness, the output text, whatever it be, is probable to have the same emotion.
\\
However, we remind you that our goal is not to reach a State-of-The-Art result but to examine this issue in Large-scale Language models.
These findings highlight the enhanced performance of our proposed AI system compared to other LLMs, even in inherently unimodal tasks. They also demonstrate the positive impact of using independent perception as generated images.

\section{Conclusion and Future Outlook}

In this paper, we introduced an imagination-inspired module to AI systems for the first time, drawing inspiration from philosophy and psychoanalysis. We incorporated a text-to-image model, enabling a process similar to ``Re-construction of Experiences". This innovation allows AI systems to develop unique interpretations that can significantly deviate from human interpretations, marking the advent of AI's ``Interpretable Misunderstanding".
In the zero-shot approach, our system outperformed the best Large Language Models (LLM) in emotion recognition tasks on the MELD, IEMOCAP, and question-answering tasks on CoQA datasets.
\\
This is just the beginning. Looking forward, we envision more intuitive AI systems. For instance, when told about ``walking on the beach", it could process the visualization of the grandeur sky, the tumultuous waves, and countless other details that weren't explicitly mentioned. These visualizations will not merely be numerical but interpretable for humans. We anticipate the next versions of our AI system that can generate other modalities from input in forms such as audio, and even action, to complete the scene with the sound of the ocean and the sense of walking. This approach could broaden our interaction spectrum with AI systems.

\bibliography{bybwfymnbfksmzhkrkywwhxcwxwyqfqc/arxiv}

\appendix

\section{Datasets}
    \begin{figure*}[t]
    \centering
    \includegraphics[width=0.8\textwidth]{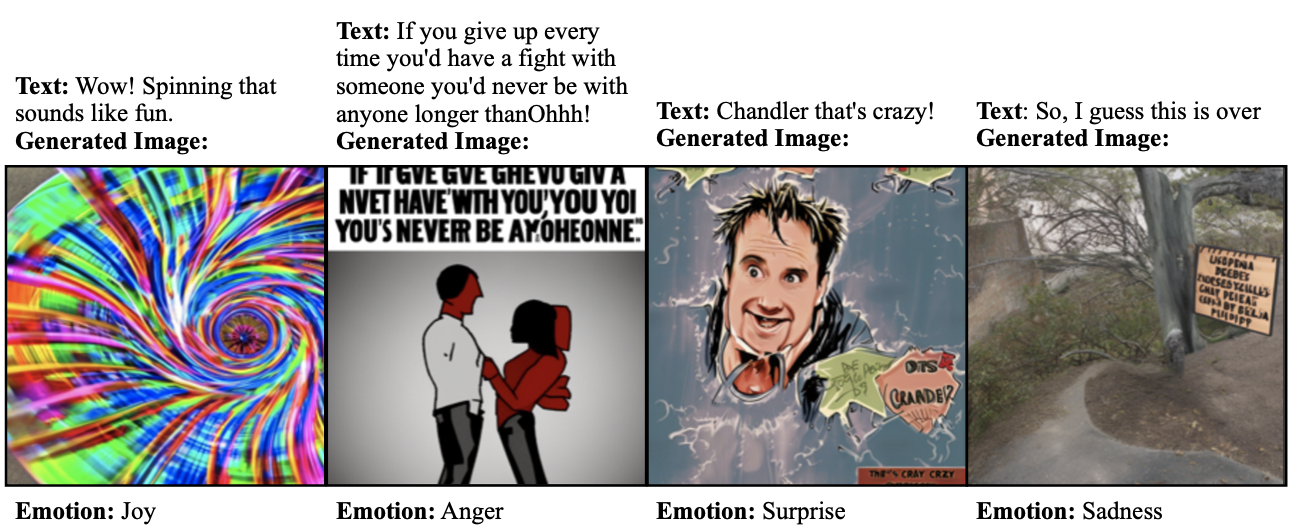}
    \caption{Example pairs of input text, generated image, and true label in MELD dataset.}
    \label{MELD_exp}
    \end{figure*}
    
    \begin{figure*}[t]
    \centering
    \includegraphics[width=0.8\textwidth]{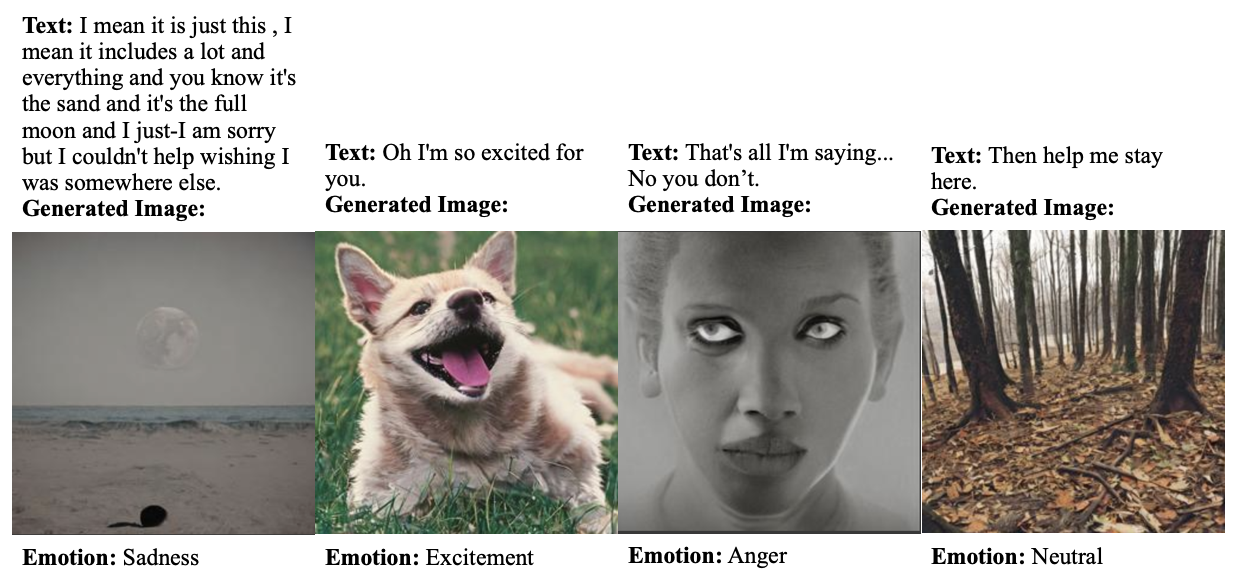}
    \caption{Example pairs of input text, generated image, and true label in the IEMOCAP dataset.}
    \label{IEMOCAP_exp}
    \end{figure*}
    
    \begin{figure*}[t]
    \centering
    \includegraphics[width=0.65\textwidth]{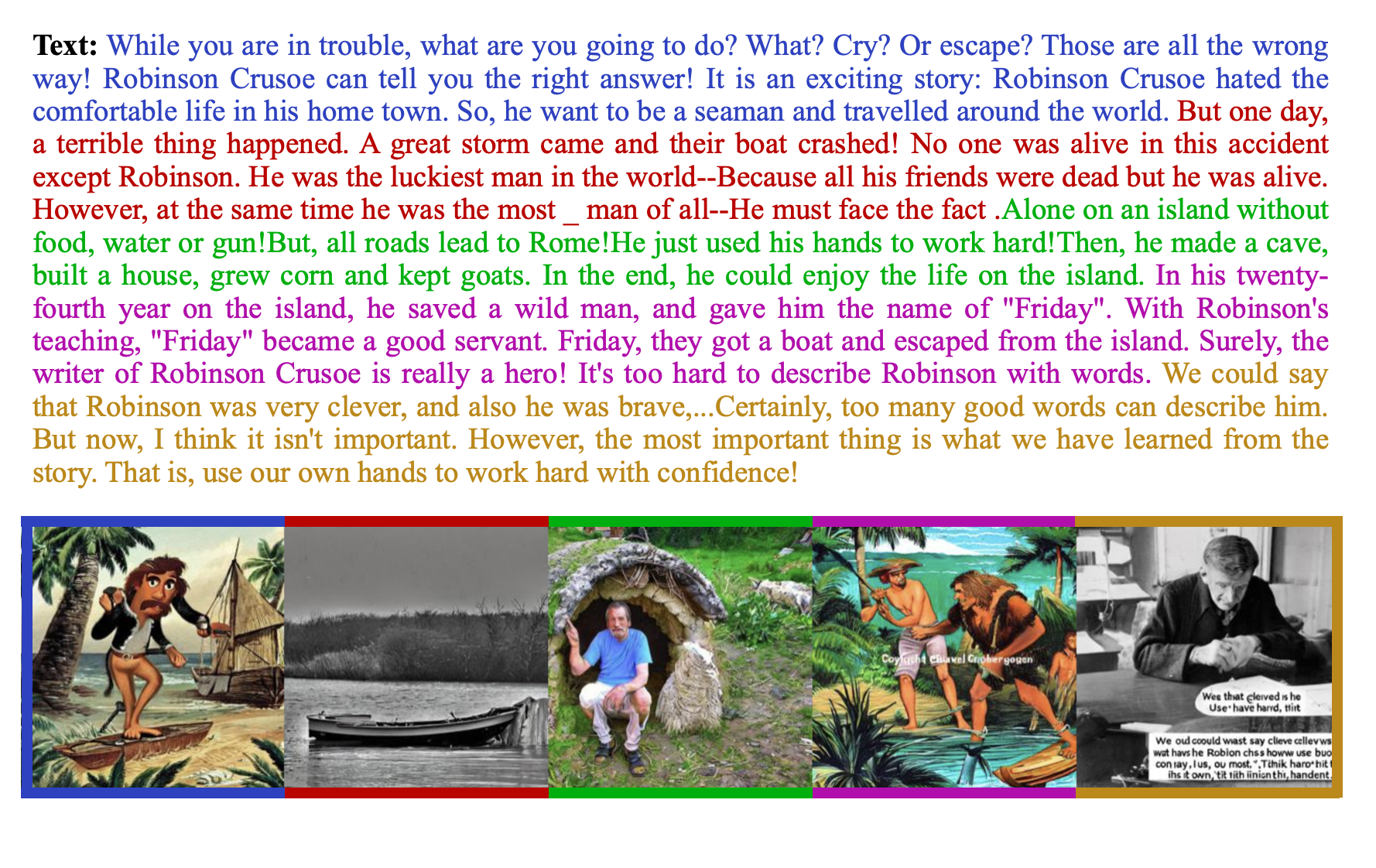} 
    \caption{\textit{An example of images produced by the text-to-image conversion model from the CoQA dataset.}
     As you can see, there is information related to the respective story in each image, and at the same time, each image is well distinguished from the other. Here, every text and image border produced from it is the same color.}
    \label{Robinson_5Pic}
    \end{figure*}
    
    \begin{figure*}[t]
    \centering
    \includegraphics[width=0.65\textwidth]{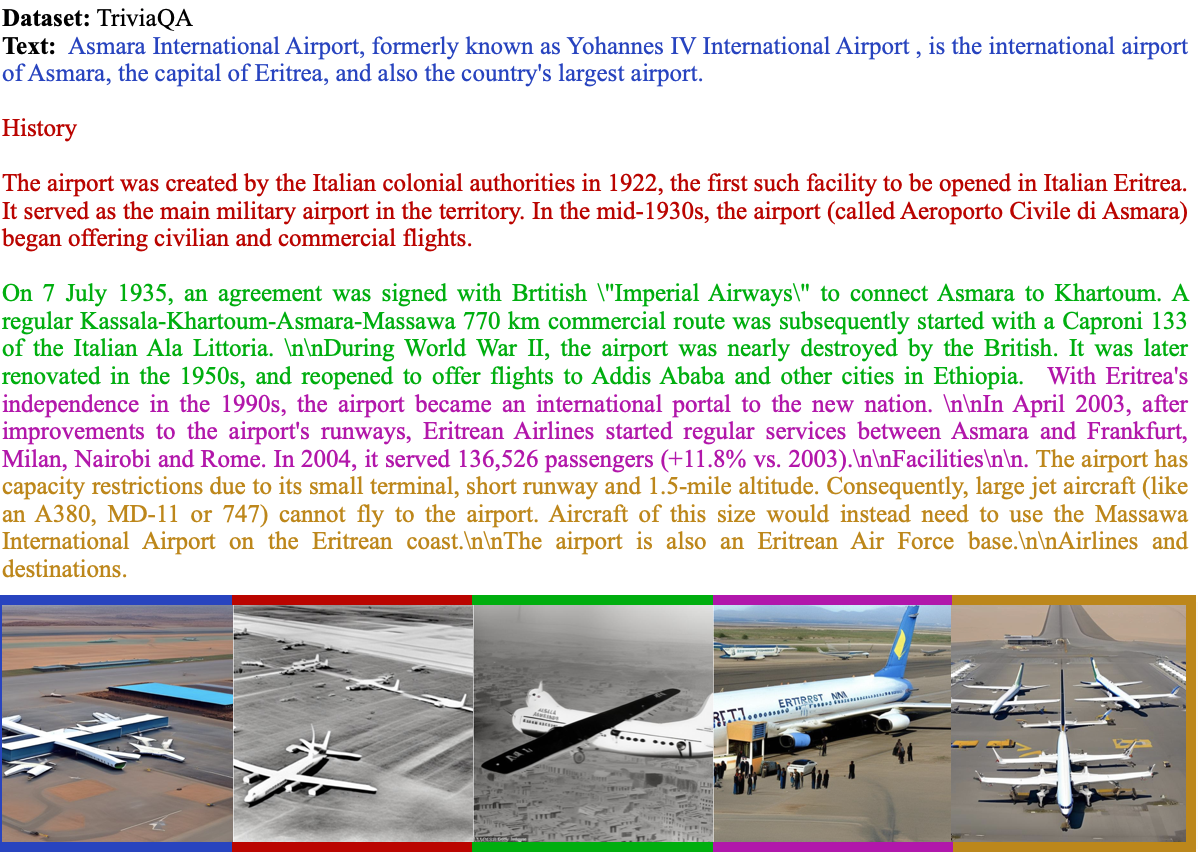} 
    \caption{\textit{An example of images produced by the text-to-image conversion model from the TriviaQA dataset.}
    This is a sample of Trivia data. We have gone through the same process for this sample as we did for the CoQA dataset.}
    \label{Trivia_QA_exp}
    \end{figure*}
    
    \begin{figure*}[t]
    \centering
    \includegraphics[width=0.65\textwidth]{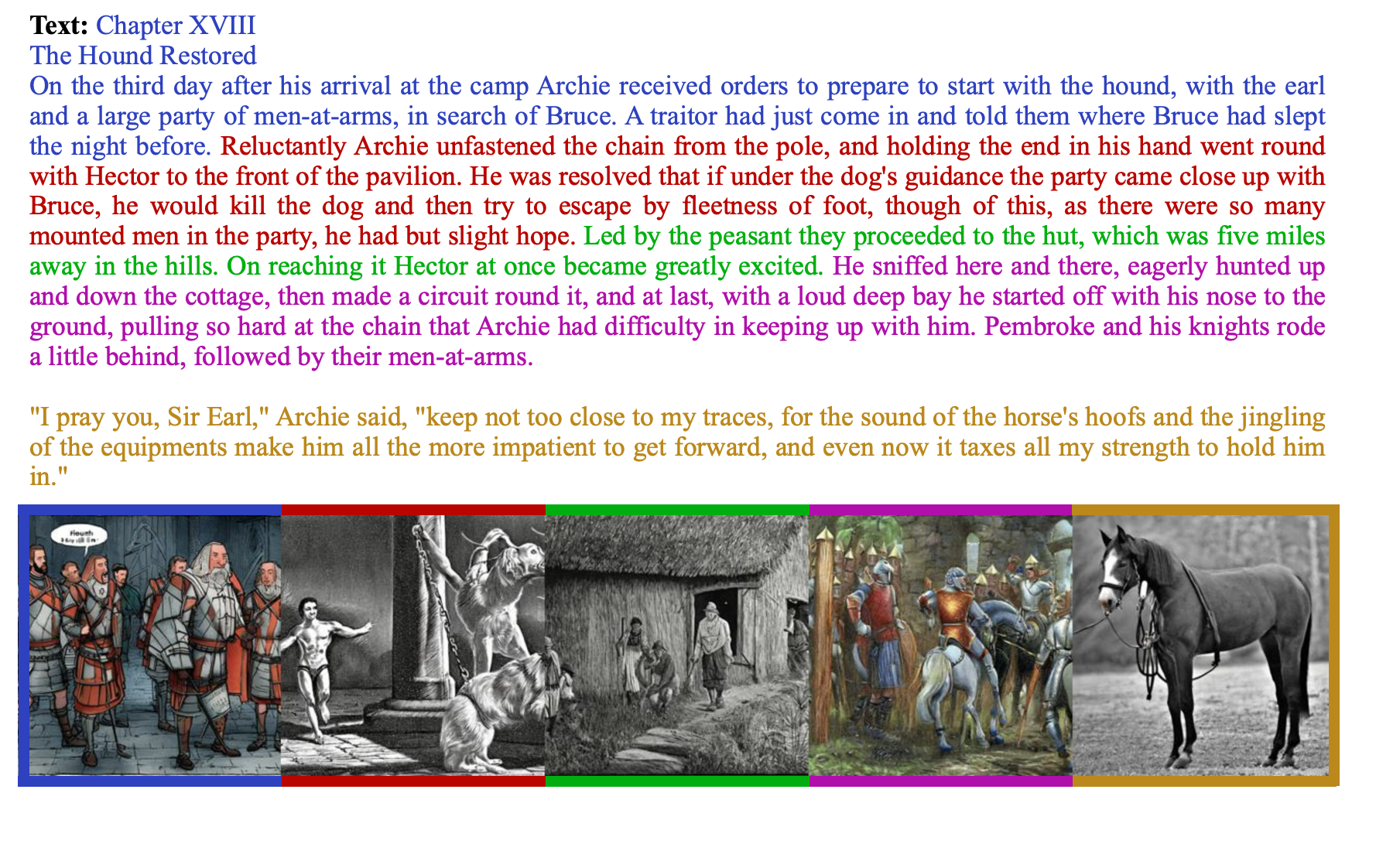} 
    \caption{\textit{Example of CoQA dataset.}
    As you can see in this example, while maintaining the general state of the story, which is about war, in each example, the main element of that part of the story is present.
}
    \label{XVIII_5Pic}
    \end{figure*}
    
    \begin{figure*}[t]
    \centering
    \includegraphics[width=0.65\textwidth]{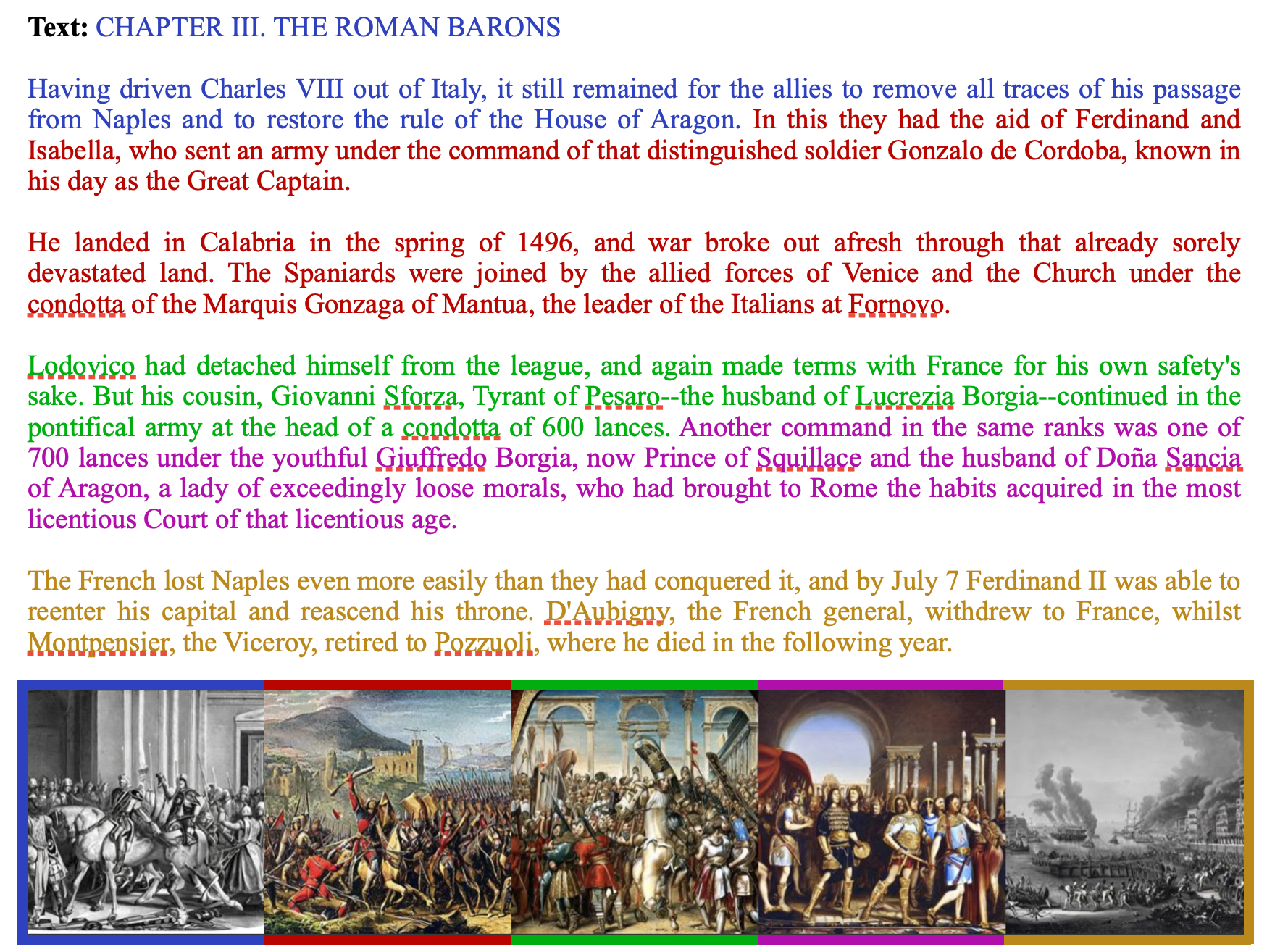} 
    \caption{
    \textit{ Another example of CoQA dataset.}
    In this example of the data, the addition of the unmentioned details can be well observed. In this way, the story has the atmosphere of "Rome" and we can well see the added details.
    From the style of the produced images to the architecture and the type of people's clothing and other things.}
    \label{ChapterIII_5Pic}
    \end{figure*}
   
    \begin{figure*}[t]
    \centering
    \includegraphics[width=0.45\textwidth]{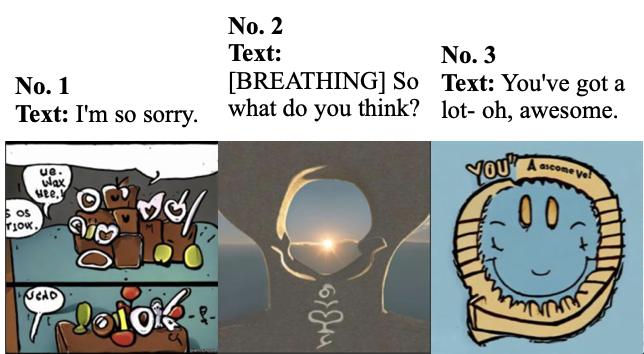}
    \caption{\textit{The input data and the corresponding generated image.}
      As you can see in the image, we created the corresponding images of the input data. This happens in the inspired measure of imagination.}
    \label{IEMOCAP_pipeline_1}
    \end{figure*}
    
    \subsection{Emotion Recognition}
    
    We have selected the MELD and IEMOCAP databases to evaluate the performance of our system in the task of emotion recognition due to the following reasons:

    \begin{itemize}
        \item Apart from textual data, these datasets also provide corresponding data in other modalities, such as audio and video. This feature enables us to assess the effectiveness of our imagination-inspired module in generating outputs that resemble real data across various modalities in future stages of our research. That is, if we ask ourselves at the next stage, ``Now that we have created this module inspired by imagination, how close are the data produced by this measure to the expected real data?'' now we have Valid data to measure the performance in comparison with them. It should be considered that this measure is beyond the ability of this module to produce natural samples. For example, if it is the modality of the image if we use the text ``How happy I am." the module generates an image, making it consistent with human logic (what text-to-image systems are trained to minimize the value of the loss function between the generated image and the actual resolution) is an issue, but We need to be able to measure how similar this generated image is to that actual situation. Or if we consider that in the future we will go to other modalities, we can use this example that the voice produced from the sentence ``I'm afraid'' is natural, but it is also important how similar the voice of the same person in the same situation is.  
        
        \item Compared to other emotion recognition datasets available in the field, these two datasets offer a diverse range of text samples that are derived from everyday conversations. They simulate ordinary dialogues among individuals, reflecting the natural interactions that occur in real-life scenarios.
    \end{itemize}
    
    You can see some pairs of input text, generated image, and corresponding true label in Figure \ref{IEMOCAP_exp}

    \subsection{Question Answering}

    We have selected the coQA dataset to evaluate the performance of our system in the task of question and answer for the following reasons:
    
    \begin{itemize}
    \item One noteworthy characteristic of this dataset is the presence of interconnected questions, where some input questions serve as continuations of previous ones. This structure allows us to explore the realm of multi-turn question and answer systems. For instance, in two consecutive questions, they may be formulated as follows: "What color was Cotton? Where did she live?" Here, the second question refers to the person mentioned in the preceding question. While our current research focuses on single-question scenarios, this dataset provides an opportunity to extend our investigations into the field of multi-turn question and answer systems.
    
    \item The coQA dataset adopts a narrative mode, which is evident in the accompanying images. These images are derived from the input text and contribute to the overall storytelling by providing meaningful visual representations. For instance, in the Robinson reference image, the pictures progressively depict the story, showcasing elements such as adventure, sailing, solitude, cultivation, and the presence of the author, as you can see in Figure \ref{Robinson_5Pic}. In contrast, other datasets like TriviaQA are primarily fact-based, as depicted in Figure \ref{Trivia_QA_exp}. The rich diversity of data in the coQA dataset enables us to explore our system's capabilities across various data types, facilitating a comprehensive evaluation of its strengths and weaknesses.
    
    \item The input text in the coQA dataset contains all the necessary information for processing. Since our objective is to assess the AI system's ability to comprehend the input text, it is crucial that the input data allows the system to answer questions based on its acquired prior knowledge (background knowledge) obtained during training. Thus, the absence of additional external knowledge in the input data ensures a focus on the system's internal understanding and reasoning abilities.
    
    \item Another benefit of these stories is that it allows us to add a nice detail to the images. For example, in the story ``CHAPTER II. ON A MOUNTAIN PATH" in Figure \ref{XVIII_5Pic}, from where it describes a war, we see that our images have a darker color and details such as armor, helmets, and other things are added. While they are not explicitly mentioned in the story. Or where he talks about Rome in the story ``CHAPTER III. THE ROMAN BARONS" in Figure \ref{ChapterIII_5Pic}, we see that our images add details of Roman architecture, or the way people dressed at that time.
    \end{itemize}
    
    \subsection{Access to Datasets}

\begin{table*}[t]
  \centering
  \begin{tabular}{|p{5cm}|p{8cm}|}
    \hline
    \textbf{Experiment} & \textbf{Instruction} \\
    \hline
    Gen\_Image\_Inp\_Text\_Both & ``What emotions do you think this pair of IMAGE and TEXT has?" + Datasets Emotions\\
    Gen\_Image\_Inp\_Text\_P1 & ``This is a classification task, choose one of the emotions." + Datasets Emotions \\
  Gen\_Image\_Inp\_Text\_P2 & ``What emotion do you perceive in this sentence" \\
  Gen\_Image\_Inp\_Text\_P3 & - \\
    \hline
  \end{tabular}
  \caption{Instructions template corresponding to the each experiment}
  \label{Experiment_Instruction}
\end{table*}

\begin{table*}[t]
    \centering
    \begin{tabular}{|p{0.5cm}|p{10.5cm}|p{2cm}|p{2cm}|}
    \hline
    \textbf{No.} & \textbf{Input Instruction} &  \textbf{Output} & \textbf{Label} \\ 
    \hline
    1 & \textbf{BEGINNING OF CONVERSATION:              USER}: what emotions do you think this \textbf{pair of IMAGE and TEXT} has?             you answer should be one of following emotions: Neutral, Happiness, Sadness, Anger, Frustration, Fear, Excitement, Disgust Surprise ,Unknown
              \textbf{TEXT} : I'm so sorry.             \textbf{Answer}: 
     & ['Sadness'] & Sadness \\ 
    \hline
    2 & \textbf{BEGINNING OF CONVERSATION:              USER}: what emotions do you think this \textbf{pair of IMAGE and TEXT} has?             you answer should be one of following emotions: Neutral, Happiness, Sadness, Anger, Frustration, Fear, Excitement, Disgust Surprise ,Unknown
              \textbf{TEXT} : [BREATHING] So what do you think?
             \textbf{Answer}: 
     & ['Unknown'] & Unknown \\ 
    \hline
    3 & \textbf{BEGINNING OF CONVERSATION:              USER}: what emotions do you think this \textbf{pair of IMAGE and TEXT} has?             you answer should be one of following emotions: Neutral, Happiness, Sadness, Anger, Frustration, Fear, Excitement, Disgust Surprise ,Unknown
              \textbf{TEXT} : You've got a lot- oh, awesome
             \textbf{Answer}: 
     & ['Excitement'] & Excitement \\ 
    \hline
    \end{tabular}
    \caption{\textbf{Experiment 1: Gen\_Image\_Inp\_Text\_Both. on IEMOCAP} This is our basic experiment. As you can see we gave these Instruction as textual inputs to out system. with the corresponding generated image. then we got MLLM's output which is not ready to be used for calculating metrics and because of that we need Helper model to give us final label. There we want MLLM to consider \textbf{both textual and visual inputs}.}
    \label{ER_results_IEMOCAP_GIB}
\end{table*}

\begin{table*}[t]
    \centering
    \begin{tabular}{|p{0.5cm}|p{10.5cm}|p{2cm}|p{2cm}|}
    \hline
    \textbf{No.} & \textbf{Input Instruction} &  \textbf{Output} & \textbf{Label} \\ 
    \hline
    1 & \textbf{BEGINNING OF CONVERSATION:              USER}: what emotions do you think this \textbf{TEXT} has?             you answer should be one of following emotions: Neutral, Happiness, Sadness, Anger, Frustration, Fear, Excitement, Disgust Surprise ,Unknown
              \textbf{TEXT} : I'm so sorry.             \textbf{Answer}: 
     & ['Sadness'] & Sadness \\ 
    \hline
    2 & \textbf{BEGINNING OF CONVERSATION:              USER}: what emotions do you think this \textbf{TEXT} has?             you answer should be one of following emotions: Neutral, Happiness, Sadness, Anger, Frustration, Fear, Excitement, Disgust Surprise ,Unknown
              \textbf{TEXT} : [BREATHING] So what do you think?
             \textbf{Answer}: 
     & ['Unknown'] & Unknown \\ 
    \hline
    3 & \textbf{BEGINNING OF CONVERSATION:              USER}: what emotions do you think this \textbf{TEXT} has?             you answer should be one of following emotions: Neutral, Happiness, Sadness, Anger, Frustration, Fear, Excitement, Disgust Surprise ,Unknown
              \textbf{TEXT} : You've got a lot- oh, awesome
             \textbf{Answer}: 
     & ['Excitement'] & Excitement \\ 
    \hline
    \end{tabular}
    \caption{\textbf{Experiment 2: Gen\_Image\_Inp\_Text\_Txt on IEMOCAP.} Here as continue of previous experiments, we gave these instructions and corresponding images to the MLLM and then used helper model to extract label. Here we want MLLM to emphasis on \textbf{textual input}.}
    \label{ER_results_IEMOCAP_GIT}
\end{table*}

\begin{table*}[t]
    \centering
    \begin{tabular}{|p{0.5cm}|p{10.5cm}|p{2cm}|p{2cm}|}
    \hline
    \textbf{No.} & \textbf{Input Instruction} &  \textbf{Output} & \textbf{Label} \\ 
    \hline
    1 & \textbf{BEGINNING OF CONVERSATION:              USER}: what emotions do you think this \textbf{IMAGE} has?             you answer should be one of following emotions: Neutral, Happiness, Sadness, Anger, Frustration, Fear, Excitement, Disgust Surprise ,Unknown
              \textbf{TEXT} : I'm so sorry.             \textbf{Answer}: 
     & ['Sadness'] & Sadness \\ 
    \hline
    2 & \textbf{BEGINNING OF CONVERSATION:              USER}: what emotions do you think this \textbf{IMAGE} has?             you answer should be one of following emotions: Neutral, Happiness, Sadness, Anger, Frustration, Fear, Excitement, Disgust Surprise ,Unknown
              \textbf{TEXT} : [BREATHING] So what do you think?
             \textbf{Answer}: 
     & ['Unknown'] & Unknown \\ 
    \hline
    3 & \textbf{BEGINNING OF CONVERSATION:              USER}: what emotions do you think this \textbf{IMAGE} has?             you answer should be one of following emotions: Neutral, Happiness, Sadness, Anger, Frustration, Fear, Excitement, Disgust Surprise ,Unknown
              \textbf{TEXT} : You've got a lot- oh, awesome
             \textbf{Answer}: 
     & ['Excitement'] & Excitement \\ 
    \hline
    \end{tabular}
    \caption{\textbf{Experiment 3: Gen\_Image\_Inp\_Text\_Img on IEMOCAP.} Here as continue of previous experiments. Instructions and corresponding images are given to the MLLM and then used helper model to extract label. Here we want MLLM to emphasis on \textbf{visual input}.}
    \label{ER_results_IEMOCAP_GII}
\end{table*}

\begin{table*}[t]
    \centering
    \begin{tabular}{|p{0.5cm}|p{10.5cm}|p{2.5cm}|p{2cm}|}
    \hline
    \textbf{No.} & \textbf{Input Instruction} &  \textbf{Output} & \textbf{Label} \\ 
    \hline
    1 & \textbf{BEGINNING OF CONVERSATION:              USER}: what emotions do you think this \textbf{IMAGE} has?             you answer should be one of following emotions: Neutral, Happiness, Sadness, Anger, Frustration, Fear, Excitement, Disgust Surprise ,Unknown
          \textbf{Answer}: 
     & ['Anger, Anger, Anger, Anger, Anger, Anger, Anger, Anger, Anger, Anger,']
 & Anger \\ 
    \hline
    2 & \textbf{BEGINNING OF CONVERSATION:              USER}: what emotions do you think this \textbf{IMAGE} has?             you answer should be one of following emotions: Neutral, Happiness, Sadness, Anger, Frustration, Fear, Excitement, Disgust Surprise ,Unknown
          \textbf{Answer}: 
     & ['Neutral, Happiness, Sadness, Anger, Frustration, Fear, Excitement, Disgust Surprise,Unknown']
 & Happiness \\ 
    \hline
    3 & \textbf{BEGINNING OF CONVERSATION:              USER}: what emotions do you think this \textbf{IMAGE} has?             you answer should be one of following emotions: Neutral, Happiness, Sadness, Anger, Frustration, Fear, Excitement, Disgust Surprise ,Unknown
          \textbf{Answer}: 
     & ['Neutral, Happiness, Sadness, Anger, Frustration, Fear, Excitement, Disgust, Surprise,Unknown']
 & Happiness \\ 
    \hline
    \end{tabular}
    \caption{\textbf{Experiment 4: Gen\_Image\_No\_Text\_Img. on IEMOCAP} As you can see we gave these Instruction without user inputs with the corresponding generated image to our MLLM to examine the richness of generated images data received from Imagination-inspired module. As you can see the outputs are longer and with variety of emotions.}
    \label{ER_results_IEMOCAP_GNI}
\end{table*}

\begin{table*}[t]
    \centering
    \begin{tabular}{|p{0.5cm}|p{10.5cm}|p{2.5cm}|p{2cm}|}
    \hline
    \textbf{No.} & \textbf{Input Instruction} &  \textbf{Output} & \textbf{Label} \\ 
    \hline
    1 & \textbf{BEGINNING OF CONVERSATION:              USER}: This is a classification Task, choose one of emotions: Neutral, Happiness, Sadness, Anger, Frustration, Fear, Excitement, Disgust Surprise ,Unknown  \textbf{TEXT}: I'm so sorry.
          \textbf{Answer}: 
     & ['Sadness']
 & Sadness \\ 
    \hline
    2 & \textbf{BEGINNING OF CONVERSATION:              USER}: This is a classification Task, choose one of emotions: Neutral, Happiness, Sadness, Anger, Frustration, Fear, Excitement, Disgust Surprise ,Unknown   \textbf{TEXT}: [BREATHING] So what do you think?
          \textbf{Answer}: 
     & ['Neutral']
 & Neutral \\ 
    \hline
    3 & \textbf{BEGINNING OF CONVERSATION:              USER}: This is a classification Task, choose one of emotions: Neutral, Happiness, Sadness, Anger, Frustration, Fear, Excitement, Disgust Surprise ,Unknown  \textbf{TEXT}: You've got a lot- oh, awesome.
          \textbf{Answer}: 
     & ['Excitement']
 & Excitement \\ 
    \hline
    \end{tabular}
    \caption{\textbf{Experiment 5: Gen\_Image\_Inp\_Text\_P1. on IEMOCAP} Here we instructed MLLM to look at the task as a classification task, not emotion recognition. Of course, we know that emotion recognition is a classification task but we wanted to examine how clearance in instructions can affect MLLM's performance.}
    \label{ER_results_IEMOCAP_GIP1}
\end{table*}

\begin{table*}[t]
    \centering
    \begin{tabular}{|p{0.5cm}|p{10.5cm}|p{2.5cm}|p{2cm}|}
    \hline
    \textbf{No.} & \textbf{Input Instruction} &  \textbf{Output} & \textbf{Label} \\ 
    \hline
    1 & \textbf{BEGINNING OF CONVERSATION:              USER}: what emotions do you perceive in one sentence ? \textbf{TEXT}: I'm so sorry.
          \textbf{Answer}: 
     & ['sadness']
 & Sadness \\ 
    \hline
    2 & \textbf{BEGINNING OF CONVERSATION:              USER}: what emotions do you perceive in one sentence ? \textbf{TEXT}: [BREATHING] So what do you think?
          \textbf{Answer}: 
     & ["I think it's a positive emotion"]
 & Happiness \\ 
    \hline
    3 & \textbf{BEGINNING OF CONVERSATION:              USER}: what emotions do you perceive in one sentence ? \textbf{TEXT}: You've got a lot- oh, awesome.
          \textbf{Answer}: 
     & ['positive']
 & Happiness \\ 
    \hline
    \end{tabular}
    \caption{\textbf{Experiment 6: Gen\_Image\_Inp\_Text\_P2 on IEMOCAP.} We gave MLLM these instructions and asked to response us the ``perceived emotion". Here MLLM was free to response us with any emotion and was not limited to dataset's label. But then by using helper model, we mapped the MLLM's output to dataset's labels.}
    \label{ER_results_IEMOCAP_GIP2}
\end{table*}

\begin{table*}[t]
    \centering
    \begin{tabular}{|p{0.5cm}|p{10.5cm}|p{2.5cm}|p{2cm}|}
    \hline
    \textbf{No.} & \textbf{Input Instruction} &  \textbf{Output} & \textbf{Label} \\ 
    \hline
    1 & I'm so sorry.
     & ["I'm so sorry i'm so sorry i'm so sorry i'm so sorry i'm so sorry "]
 & Fear \\ 
    \hline
    2 &  [BREATHING] So what do you think?
     & ['What do you think?             ']
 & Fear \\ 
    \hline
    3 & You've got a lot- oh, awesome.
     & ["You've got a lot- oh, awesome t-shirt"]
 & Excitement \\ 
    \hline
    \end{tabular}
    \caption{\textbf{Experiment 7: Gen\_Image\_Inp\_Text\_P3 on IEMOCAP.} As you can see, there is no instruction there. we just gave textual and visual inputs to the MLLM and evaluated the MLLM's response.}
    \label{ER_results_IEMOCAP_GIP3}
\end{table*}

 \begin{figure*}[t]
    \centering
    \includegraphics[width=0.5\textwidth]{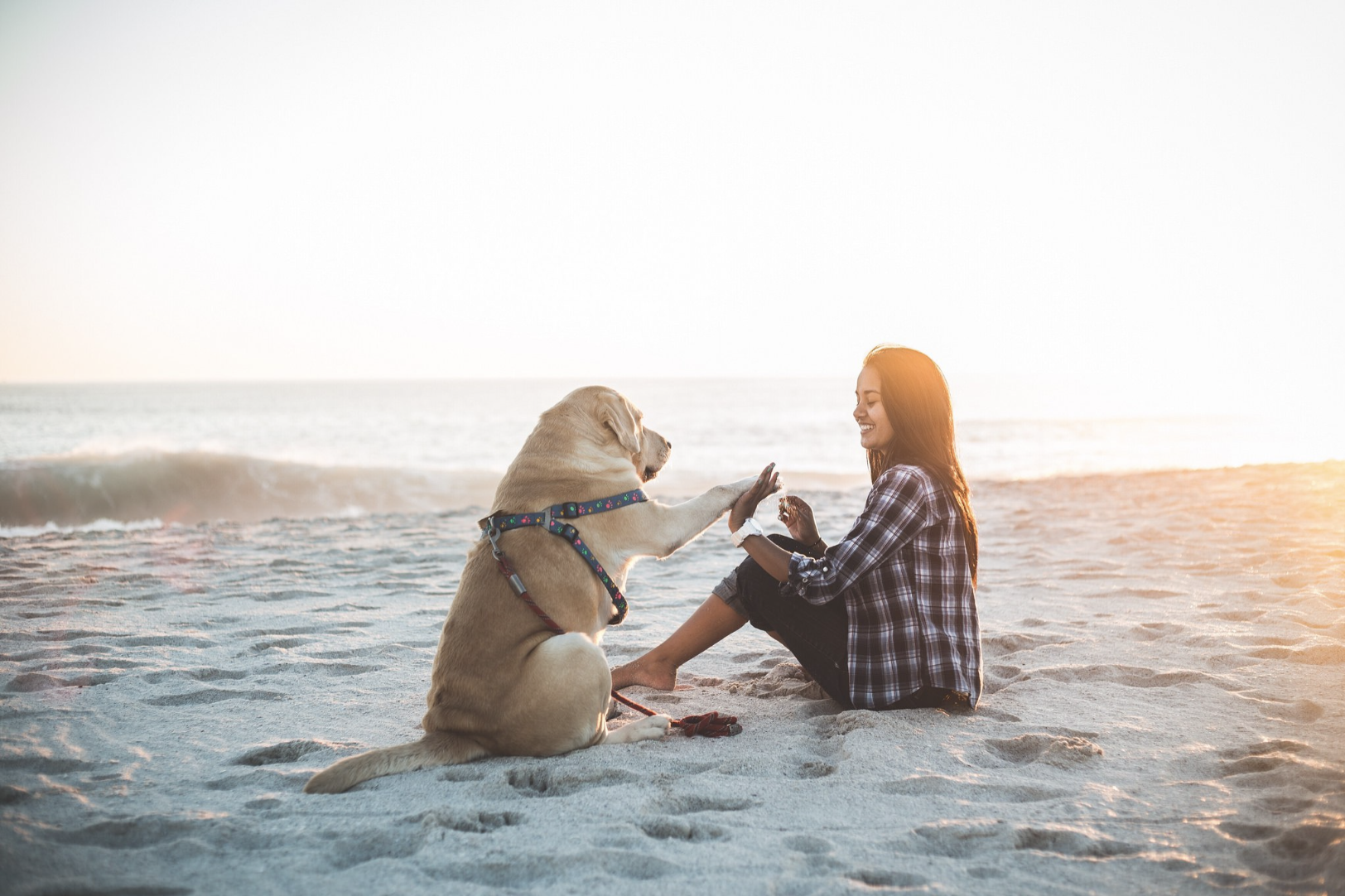}
    \caption{Demo image provided in instructblip GitHub. We used it as constant image to examine effect of using generated image in comparison with.}
    \label{Demo_img}
    \end{figure*}

\begin{table*}[t]
    \centering
    \begin{tabular}{|p{0.5cm}|p{10.5cm}|p{2.5cm}|p{2cm}|}
    \hline
    \textbf{No.} & \textbf{Input Instruction} &  \textbf{Output} & \textbf{Label} \\ 
    \hline
    1 & \textbf{BEGINNING OF CONVERSATION:              USER}: what emotions do you think pair of IMAGE and TEXT has?             you answer should be one of following emotions: Neutral, Happiness, Sadness, Anger, Frustration, Fear, Excitement, Disgust Surprise ,Unknown \textbf{TEXT}: [BREATHING] So what do you think?
          \textbf{Answer}: 
     & ['Sadness']
 & Sadness \\ 
    \hline
    2 &  \textbf{BEGINNING OF CONVERSATION:              USER}: what emotions do you think pair of IMAGE and TEXT has?             you answer should be one of following emotions: Neutral, Happiness, Sadness, Anger, Frustration, Fear, Excitement, Disgust Surprise ,Unknown \textbf{TEXT}: [BREATHING] So what do you think?
          \textbf{Answer}: 
     & ['Unknown']
 & Unknown \\ 
    \hline
    3 & \textbf{BEGINNING OF CONVERSATION:              USER}: what emotions do you think pair of IMAGE and TEXT has?             you answer should be one of following emotions: Neutral, Happiness, Sadness, Anger, Frustration, Fear, Excitement, Disgust Surprise ,Unknown \textbf{TEXT}: [BREATHING] So what do you think?
          \textbf{Answer}: 
     & ['Excitement']
 & Excitement \\ 
    \hline
    \end{tabular}
    \caption{\textbf{Experiment 8: Dem\_Image\_Inp\_Text\_Both. on IEMOCAP} As you can see we gave these Instruction with user inputs and the Demo image to our MLLM.}
    \label{ER_results_IEMOCAP_DIB}
\end{table*}

\begin{table*}[t]
    \centering
    \begin{tabular}{|p{0.5cm}|p{6.5cm}|p{6.5cm}|p{2cm}|}
    \hline
    \textbf{No.} & \textbf{Input Instruction} &  \textbf{LLM's Answer} & \textbf{Label} \\ 
    \hline
    1 & \textbf{BEGINNING OF CONVERSATION:              USER}: what emotions do you think this \textbf{Text} has?             you answer should be one of following emotions: Neutral, Happiness, Sadness, Anger, Frustration, Fear, Excitement, Disgust Surprise ,Unknown.
    \textbf{Text}: I'm so sorry.

          \textbf{Answer}: 
     &    
     "BEGINNING OF CONVERSATION:              USER : what emotion do you think this TEXT has?              you answer should be one of following emotions: Neutral, Happiness, Sadness, Anger, Frustration, Fear, Excitement,Disgust,Surprise,Unknown             TEXT :  I'm so sorry.             Answer:   "
    & Anger 
    \\ 
    \hline
    2 & \textbf{BEGINNING OF CONVERSATION:              USER}: what emotions do you think this \textbf{TEXT} has?             you answer should be one of following emotions: Neutral, Happiness, Sadness, Anger, Frustration, Fear, Excitement, Disgust Surprise ,Unknown.
    \textbf{Text}: [BREATHING] So what do you think? 
          \textbf{Answer}: 
     &   'BEGINNING OF CONVERSATION:              USER : what emotion do you think this TEXT has?              you answer should be one of following emotions: Neutral, Happiness, Sadness, Anger, Frustration, Fear, Excitement,Disgust,Surprise,Unknown             TEXT :  [BREATHING] So what do you think?             Answer:  Happiness'

 & Neutral \\ 
    \hline
    3 & \textbf{BEGINNING OF CONVERSATION:              USER}: what emotions do you think this \textbf{TEXT} has?             you answer should be one of following emotions: Neutral, Happiness, Sadness, Anger, Frustration, Fear, Excitement, Disgust Surprise ,Unknown.
    \textbf{Text}: You've got a lot- oh, awesome.   
          \textbf{Answer}: 
     & 'BEGINNING OF CONVERSATION:              USER : what emotion do you think this TEXT has?              you answer should be one of following emotions: Neutral, Happiness, Sadness, Anger, Frustration, Fear, Excitement,Disgust,Surprise,Unknown             TEXT :  You've got a lot- oh, awesome.             Answer: Surprised
     
 & Excitement \\ 
    \hline
    \end{tabular}
    \caption{\textbf{Experiment 9: Koala-7B on IEMOCAP.} We gave these Instructions to LLM. then we got LLM's output which as you can see, contains input at beginning and makes helper model to choose different label as what was responded as answer.}
    \label{ER_results_IEMOCAP_Koala}
\end{table*}

\begin{table*}[t]
    \centering
    \begin{tabular}{|p{0.5cm}|p{10.5cm}|p{2.5cm}|p{2cm}|}
    \hline
    \textbf{No.} & \textbf{Input Instruction} &  \textbf{Extracted Answer} & \textbf{Label} \\ 
    \hline
    1 & \textbf{BEGINNING OF CONVERSATION:              USER}: what emotions do you think this \textbf{Text} has?             you answer should be one of following emotions: Neutral, Happiness, Sadness, Anger, Frustration, Fear, Excitement, Disgust Surprise ,Unknown.
    \textbf{Text}: I'm so sorry.

          \textbf{Answer}: 
     &    [{" "}]
 & Disgust \\ 
    \hline
    2 & \textbf{BEGINNING OF CONVERSATION:              USER}: what emotions do you think this \textbf{TEXT} has?             you answer should be one of following emotions: Neutral, Happiness, Sadness, Anger, Frustration, Fear, Excitement, Disgust Surprise ,Unknown.
    \textbf{Text}: [BREATHING] So what do you think? 
          \textbf{Answer}: 
     &   [{'Happiness'}]
 & Happiness \\ 
    \hline
    3 & \textbf{BEGINNING OF CONVERSATION:              USER}: what emotions do you think this \textbf{TEXT} has?             you answer should be one of following emotions: Neutral, Happiness, Sadness, Anger, Frustration, Fear, Excitement, Disgust Surprise ,Unknown.
    \textbf{Text}: You've got a lot- oh, awesome.   
          \textbf{Answer}: 
     & [{"Surprised"}]
 & Surprise \\ 
    \hline
    \end{tabular}
    \caption{\textbf{Experiment 10: Koala-7B with ``output-processing" on IEMOCAP.} We gave these instructions to MLLM with images. there you can see extracted answer from MLLM's output and the corresponding label.}
    \label{ER_results_IEMOCAP_Koala_pr}
\end{table*}

	In this section, to enhance clarity and facilitate a comprehensive understanding of the article's process, we provide sample inputs and outputs from a series of tests. The results for the emotion recognition task are presented for the IEMOCAP dataset:
	
	\subsection{Our AI System Experiments}
	
	Here we present all out experiments. Based on our system's architecture, at first we should converts texts to images. These images are the same for all experiment.
	you can see inputs and corresponding generated image in Figure \ref{IEMOCAP_pipeline_1}
	
	There are experiments there: 

	\begin{itemize}
	
    \item  Gen\_Image\_Inp\_Text\_Both : Table \ref{ER_results_IEMOCAP_GIB}
    
    \item Gen\_Image\_Inp\_Text\_Txt : Table \ref{ER_results_IEMOCAP_GIT}
    
    \item Gen\_Image\_Inp\_Text\_Img : Table \ref{ER_results_IEMOCAP_GII}
    
    \item Gen\_Image\_Np\_Text\_Img : Table \ref{ER_results_IEMOCAP_GNI}
    
    \item Gen\_Image\_Inp\_Text\_P1: Table \ref{ER_results_IEMOCAP_GIP1}
    
    \item Gen\_Image\_Inp\_Text\_P2 : Table \ref{ER_results_IEMOCAP_GIP2}
    
    \item Gen\_Image\_Inp\_Text\_P3 : Table \ref{ER_results_IEMOCAP_GIP3}
    
    \item Dem\_Image\_Inp\_Text\_Both : Table \ref{ER_results_IEMOCAP_DIB}. The Demo image is provided in instructblip GitHub\footnote{https://github.com/salesforce/LAVIS/tree/main/projects/
    \\instructblip} as you can see in Figure \ref{Demo_img}.
    
\end{itemize}

    For experiment P1, P2 and P3, the exact instructions are provided in Table \ref{Experiment_Instruction}.

	\subsection{Best LMM Experiment}
	
	In the Emotion Recognition task, the best model was Koala-7B, but as it was told before, the primary output of the model contains all input at the beginning, and for that, we extracted the answer field and examined the new results as ``output-processing".
	So here we have two experiments and results:
	
	\begin{itemize}
	    \item Koala-7B : Table \ref{ER_results_IEMOCAP_Koala}
	    \item Koala-7B with output-processing : Table \ref{ER_results_IEMOCAP_Koala_pr}
	\end{itemize}
\section{Implementation Tips}
 The tips for implementing the experiment are mentioned below. All tests were performed on an A100, and the time for each test was less than 4 hours. For ease of work and optimal use of processing resources, the part of producing images from the text is done in a separate part, and the images are saved, and in fact, in part of the process, we also produced a dataset. Finally, the pair of related images and text is given to MLLM.

\end{document}